\documentclass{article}
\usepackage[preprint]{spconf,amsmath,graphicx}
\copyrightnotice{\copyright\ IEEE 2022}
\toappear{To appear in {\it Proc.\ ICASSP2022, May 22-27, 2022, Singapore, Singapore}}
\usepackage{amsfonts}
\usepackage{hyperref}
\usepackage{bm}
\usepackage{float}
\usepackage{nicefrac}
\usepackage{support-caption}
\usepackage{subcaption}
\usepackage{multirow}
\usepackage{multicol}
\usepackage{makecell}
\usepackage{adjustbox}
\usepackage{booktabs}
\usepackage{tablefootnote}
\usepackage{xspace}

\def\fairseq{\textsc{fairseq}\xspace}

\title{Wav2vec-Switch: Contrastive Learning from Original-noisy Speech Pairs for Robust Speech Recognition}
\name{Yiming Wang$^\dagger$ \qquad Jinyu Li$^\dagger$ \qquad Heming Wang$^\star$\sthanks{Work done during an internship at Microsoft.} \qquad Yao Qian$^\dagger$ \qquad Chengyi Wang$^\dagger$ \qquad Yu Wu$^\dagger$}
\address{
  $^\dagger$ Microsoft Corporation \qquad $^\star$ The Ohio State University, USA \\
\small{\texttt{\{yimingwang,jinyli,yaoqian,t-chewang,wu.yu\}@microsoft.com, wang.11401@osu.edu}}
}
\begin{document}
\ninept
\maketitle
\begin{abstract}
The goal of self-supervised learning (SSL) for automatic speech recognition (ASR) is to learn good speech representations from a large amount of unlabeled speech for the downstream ASR task. However, most SSL frameworks do not consider noise robustness which is crucial for real-world applications. In this paper we propose wav2vec-Switch, a method to encode noise robustness into contextualized representations of speech via contrastive learning. Specifically, we feed original-noisy speech pairs simultaneously into the wav2vec 2.0 network. In addition to the existing contrastive learning task, we switch the quantized representations of the original and noisy speech as additional prediction targets of each other. By doing this, it enforces the network to have consistent predictions for the original and noisy speech, thus allows to learn contextualized representation with noise robustness. Our experiments on synthesized and real noisy data show the effectiveness of our method: it achieves 2.9--4.9\% relative word error rate (WER) reduction on the synthesized noisy LibriSpeech data without deterioration on the original data, and 5.7\% on CHiME-4 real 1-channel noisy data compared to a data augmentation baseline even with a strong language model for decoding. Our results on CHiME-4 can match or even surpass those with well-designed speech enhancement components.
\end{abstract}
\section{Introduction}
\label{sec:intro}
Training an automatic speech recognition (ASR) system without the need to collect a large amount of transcribed data has been a long-lasting problem, especially for low-resource domains/languages. Previous efforts include model transfer learning, domain adaptation, knowledge distillation/teacher-student learning, and semi-supervised training  \cite[\emph{inter alia}]{wang2015transfer,das2015cross,asami2017domain,li2017large,wang2020improving,manohar2018semi}. Recently, self-supervised learning (SSL) has emerged as a promising paradigm to tackle this problem. SSL for speech tasks leverages unlabeled data with a self-supervised loss in a pre-training stage, where it is capable of learning good contextualized representations from input speech. Then after fine-tuning the pre-trained model with a small amount of transcribed speech in a conventional supervised manner, the performance can match those trained directly with a much larger amount of labeled data \cite{oord2018representation,schneider2019wav2vec,chung2020generative,liu2020mockingjay,ling2020deep,wang2021unispeech,hsu2021hubert}. Existing SSL methods for speech include contrastive predictive coding (CPC) \cite{oord2018representation,schneider2019wav2vec}, auto-regressive predictive coding \cite{chung2020generative}, and masked predictive encoding \cite{liu2020mockingjay,baevski2020vq,hsu2021hubert}. Certain others may fit in more than one categories above \cite{baevski2020wav2vec,chung2021w2v}. Moreover, a recent work \cite{zhang2020pushing} iteratively performed pseudo-labeling in a more traditional way and fine-tuning on refined pre-trained models to further push the limits of SSL. 

Noise robustness is another challenge for ASR in real-world applications \cite{li2015robust}. Speech recordings from real-world scenarios usually contain background noise and noise caused by recording imperfection, resulting in deteriorated ASR performance. Prevailing strategies dealing with this challenge are to plug a dedicated enhancement/denoising module into the pipeline of an ASR system as a frond-end to suppress the noise, either by training that module separately \cite{vincent2017analysis,kinoshita2020improving,wang2020complex} or jointly \cite{subramanian2019speech,chang2019mino} with acoustic models. The motivation of joint training is to alleviate the problem that optimizing enhancement objective independently does not necessarily lead to an optimal solution for the ASR task, even if it improves the intelligibility of speech. In either way it will add complexity to the neural network models.

In this paper we focus on strengthening the noise robustness of the pre-trained model during SSL for ASR. Existing work to this end includes PASE+ \cite{ravanelli2020multi} where a variety of speech transformations were estimated from contaminated speech, and wav2vec-C \cite{sadhu2021wav2vec} where a reconstruction module was added on top of the quantized output of the wav2vec 2.0 network \cite{baevski2020wav2vec} and the reconstruction loss was jointly trained with the existing contrastive loss. Same as wav2vec-C, our work, wav2vec-Switch, is also based on wav2vec 2.0. However, instead of a reconstruction loss, we add another contrastive loss as an auxiliary task to achieve noise robustness. Specifically, we batch both the original speech\footnote{We refer to ``original speech'' rather than ``clean speech'' to avoid any possible confusion, as the original speech in our case is not necessarily clean.} and its noisy version together and feed them to the network. Then in the contrastive learning the quantized targets in each original-noisy pair are switched, so that both the targets are treated as positive in their respective loss calculation. The motivation is that, if we want the contextualized representation robust to noise, the representation of an original speech should also be able to predict the target of its noisy version and vice versa. Different from the prior work, ours does not involve a process of transforming from one representation to another, but enforces the prediction consistency constraint in the contrastive loss without adding any complexity to networks. Experiments on synthesized noisy data and real-world noisy data from the CHiME-4 challenge \cite{vincent2017analysis} indicate the efficacy of our approach. In particular, compared to the baseline with data augmentation applied only, we observe 7.1--11.0\% relative word error rate (WER) reduction on synthesized noisy speech and 7.8\% reduction on real-world noisy speech when decoding without language model (LM), while still maintaining the performance on the original speech. Even in the presence of a strong neural LM, the WER reduction is 2.9--4.9\% and 5.7\% respectively. Moreover, our results on the CHiME-4 data, with only the unlabeled 960-hour LibriSpeech audio for pre-training, are comparable to, or even better than, other work with a complicated speech enhancement module.

\section{Models}
\vspace{-1mm}

\subsection{Wav2vec 2.0}
\vspace{-1mm}
\label{sec:wav2vec2}
We recapitulate the wav2vec 2.0 model on which our method is based. Compared to its precedents \cite{schneider2019wav2vec,baevski2020vq}, wav2vec 2.0 combines masked prediction and contranstive learning into a unified model during pre-training. It has a feature encoder $f: \mathcal{X} \mapsto \mathcal{Z}$ with a raw audio waveform $\mathbf{x} \in \mathbb{R}^T$ as its input, and a latent representation $Z=[\mathbf{z}_1,\ldots,\mathbf{z}_{T'}]$ with a time-domain down-sampling through a set of convolutional blocks as its output; a context network $g: \mathcal{Z} \mapsto \mathcal{C}$ that takes the masked $Z$ and outputs a contextualized representation $\mathbf{c}_t$ at each masked position $t$ through several blocks of Transformers \cite{vaswani2017attention}; and a quantization module $h: \mathcal{Z} \mapsto \mathcal{Q}$ discretizing the unmasked $Z$ to $Q$ from a finite set of codebook via Gumbel Softmax \cite{jang2017categorical} and product quantization \cite{jegou2011product}. The contrastive loss is applied at each masked position $t$, discriminating the true quantized representation $\mathbf{q}_t$ (the positive sample) from $K$ distractors $Q_t^-=\{\mathbf{q}^-_1,\ldots,\mathbf{q}^-_K\}$ (the negative samples) drawn from other masked positions within the same training example:
\begin{eqnarray}
\label{eq:contrastive_loss}
\mathcal{L}^\mathrm{C}(C,Q) &=& \sum_{t=1}^N \mathcal{L}_t^\mathrm{C}(C,Q)/N \\
\mathcal{L}_t^\mathrm{C}(C,Q) &=& - \log \frac{\exp(\mathrm{sim}(\mathbf{c}_t,\mathbf{q}_t))}{\sum_{\mathbf{q}^- \in Q_t^-} \exp(\mathrm{sim}(\mathbf{c}_t,\mathbf{q}^-))}
\end{eqnarray}
where $N$ is the number of masked positions where the loss is computed, and $\mathrm{sim}(\cdot,\cdot)$ is implemented as the \emph{cosine similarity} function. In addition, a diversity loss $\mathcal{L}^\mathrm{D}$ encouraging the codebook utilization is implemented as negative perplexity of the Gumbel Softmax output. The total loss to optimize is:
\begin{equation}
\label{eq:total_loss}
\mathcal{L} = \mathcal{L}^\mathrm{C} + \alpha \mathcal{L}^\mathrm{D}
\end{equation}
where $\alpha$ is a coefficient. During fine-tuning, the quantization module is discarded and the parameters inside the feature encoder are frozen. All the other network parameters are updated with the CTC loss \cite{graves2006connectionist}.

\vspace{-1mm}
\subsection{Wav2vec-Switch}
In the pre-training stage of wav2vec 2.0, the decision that distractors are only sampled from masked positions within the same training example, rather than from any examples, is critical, or it will hurt the ASR performance \cite{baevski2020wav2vec}. The reason is that sampling only within the same training example can avoid learning features irrelevant to ASR, e.g., speaker or environmental characteristics. However, no mechanisms are designed to achieve noise robustness during the pre-training: when a noisy utterance is given, both the positive and negative samples that the contrastive loss is using contain noise, and there is no explicit way to differentiate the noise from the speech or to learn a contextualized representation invariant to noise. Our intuition is that, if the contextualized representation is robust to noise, the representation of an original/noisy speech should be capable of predicting the target of the noisy/original speech as well. Motivated by this, we propose wav2vec-Switch as follows.

For each mini-batch of the original waveform $X \in \mathbb{R}^{B \times T}$ where $B$ is the batch size, we duplicate $X$, apply an independently sampled noise to each row (example), and thus have a noisy version of $X$ as $\widetilde{X}$. Then both $X$ and $\widetilde{X}$ are fed into the wav2vec 2.0 network in parallel and forwarded through the feature encoder $f$, the context network $g$, and the quantization module $h$. At this point we have 4 quantities\footnote{Here we slightly abuse the use of the notation $C,Q,\widetilde{C},\widetilde{Q}$ to denote the batched versions of their respective quantities described in Section \ref{sec:wav2vec2}.}:
\begin{eqnarray}
\label{eq:four_quantities}
C=g(f(X)), \qquad Q=h(f(X)) \\\nonumber
\widetilde{C}=g(f(\widetilde{X})), \qquad \widetilde{Q}=h(f(\widetilde{X}))
\end{eqnarray}

\begin{figure*}[ht]
\centering
\includegraphics[width=0.8\textwidth]{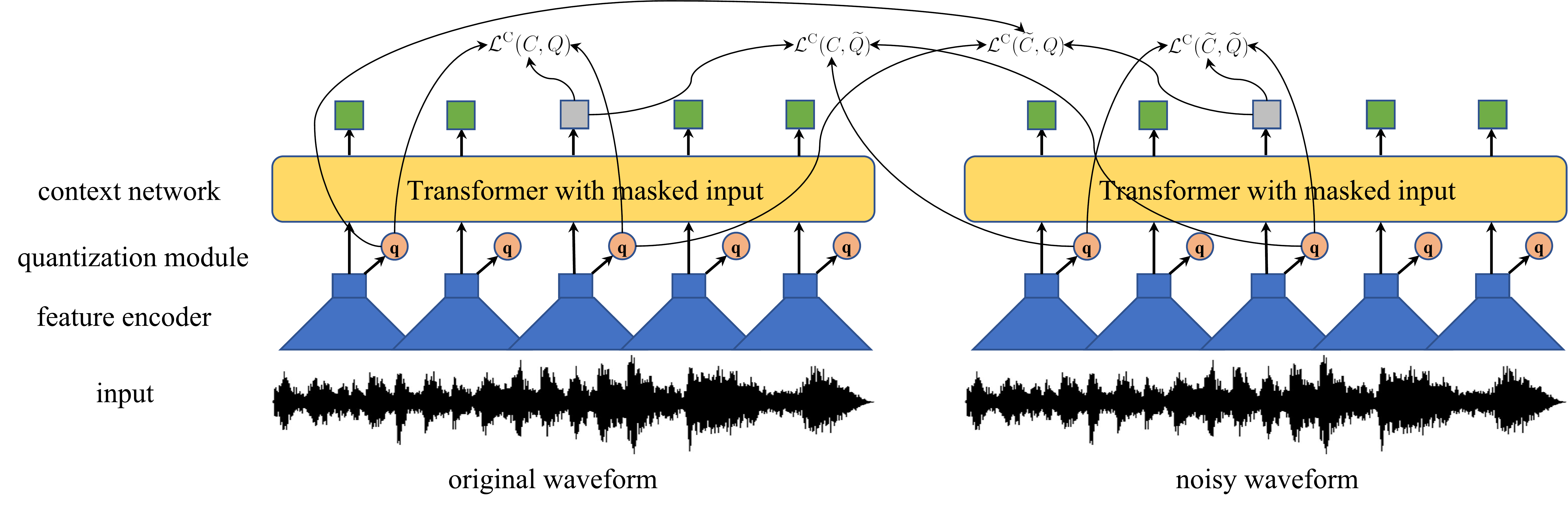}
\caption{This figure illustrates how the 4 contrastive losses are calculated in the proposed wav2vec-Switch model. The original-noisy pair of speech is fed into the network with identical internal states simultaneously. The loss quantities $\mathcal{L}^\mathrm{C}(C,\widetilde{Q})$ and $\mathcal{L}^\mathrm{C}(\widetilde{C},Q)$ in the middle are obtained by switching the quantized representations of the original and noisy speech as prediction targets.}
\label{fig:model}
\end{figure*}

In addition to the standard contrastive loss described in Eq. (\ref{eq:contrastive_loss}) where the loss takes $(C,Q)$ or $(\widetilde{C},\widetilde{Q})$ as its input arguments, we also switch the quantized targets $Q$ and $\widetilde{Q}$, and form two more tuples for the loss: $(C,\widetilde{Q})$ and $(\widetilde{C},Q)$. Therefore we obtain 4 contrastive loss quantities: $\mathcal{L}^\mathrm{C}(C,Q)$, $\mathcal{L}^\mathrm{C}(\widetilde{C},\widetilde{Q})$, $\mathcal{L}^\mathrm{C}(C,\widetilde{Q})$, and $\mathcal{L}^\mathrm{C}(\widetilde{C},Q)$. The new loss would be:
\begin{eqnarray}
\label{eq:new_contrastive_loss}
\mathcal{L}_{\mathrm{switch}}^\mathrm{C}(C,Q,\widetilde{C},\widetilde{Q})&=&\mathcal{L}^\mathrm{C}(C,Q)+\mathcal{L}^\mathrm{C}(\widetilde{C},\widetilde{Q}) + \\\nonumber &&\lambda\left(\mathcal{L}^\mathrm{C}(C,\widetilde{Q})+\mathcal{L}^\mathrm{C}(\widetilde{C},Q)\right)
\end{eqnarray}
where the coefficient $\lambda$ controls the weight of the term calculated from the switched targets, relative to the one obtained from the original targets. Fig. \ref{fig:model} also illustrates how the 4 contrastive loss quantities are calculated. Finally $\mathcal{L}^\mathrm{D}$ is added to the total loss the same way as in Eq. (\ref{eq:total_loss}). Note that when $\lambda=0$, the method reduces to the ``wav2vec 2.0 with data augmentation'' case, which will be compared against as the baselines in Section \ref{sec:exp}.

We also need to keep in mind that the network's internal states for any specific input pair $(X,\widetilde{X})$ should be \emph{identical}. \emph{Identical} means that not only the architecture and parameters but also anything inside the network that relies on random states, including masked positions for the context network and dropout masks of all the dropout layers, should be the same for $X$ and $\widetilde{X}$. Otherwise the representations of the original speech and its noisy version will not follow with each other, and our approach will not behave as what we expect or learn representations with a meaningful interpretation. The ablation study in Section \ref{sec:ablation} also verifies the importance of enforcing such a constraint.

In practice, we batch $X$ and $\widetilde{X}$ together, and feed the resulting ``large'' mini-batch to the network. Every time before a random function is invoked, we save the current random state. That random state is restored immediately after the random function is invoked for the first half of the mini-batch, and then the same function is executed for the second half. By doing in this way, we are able to ensure that the network's internal states for $X$ and $\widetilde{X}$ are always identical.

\section{Experiments}
\label{sec:exp}
\vspace{-1mm}
\subsection{Datasets}
\label{sec:datasets}
In order to validate the noise robustness of our proposed approach for ASR, we evaluate it on both synthesized and real-world noisy data. In this section, we first introduce how our data was prepared for these experiments. For the SSL paradigm, there are normally 3 stages: 1) self-supervised pre-training with a large unlabeled dataset; 2) fine-tuning with a relatively small set of labeled data; and 3) testing data in the target domain. We will explain the corresponding setup for each of these 3 stages.

In the synthesized data experiments, both the pre-training and test sets were formed by mixing the LibriSpeech corpus \cite{panayotov2015librispeech} with noise randomly drawn from the MUSAN corpus \cite{snyder2015musan}, and the signal-to-noise ratio (SNR) was uniformly sampled from the range between 5 and 10 dB. Note that the MUSAN corpus contains 3 categories of noise: \emph{music}, \emph{noise} and \emph{speech}. To avoid introducing potential confusion with actual speech when learning speech representations, we only used the first 2 categories of the noise from MUSAN and added them on-the-fly to the original 960-hour LibriSpeech training set for pre-training. For the test sets, we prepared the original \texttt{test-clean} and \texttt{test-other}, and several synthesized versions with different levels of SNR or different subsets of the MUSAN corpus being added, in order to evaluate the model on test sets with a variety of mismatched conditions. We used the original LibriSpeech \texttt{train-clean-100} labeled data for fine-tuning, and the reason is that we would like the final model not to degrade its performance on the original sets.

In the real noisy data experiments, we used the data from the CHiME-4 challenge\footnote{\url{http://spandh.dcs.shef.ac.uk/chime_challenge/CHiME4/index.html}} \cite{vincent2017analysis}. The data in CHiME-4 is based on the ``WSJ0'' portion of the Wall Streat Journal corpus \cite{paul1992design} (WSJ), either obtained by recording the actual talker's speech in real noisy environments (on a bus, cafe, pedestrian area, and street junction) (denoted as ``real data''), or artificially mixing the clean WSJ speech with background noise and recording them using a 6-channel distant microphone array and a close-talk microphone (denoted as ``simulated data''). Specifically, we chose the real 1-channel track for testing. All the channels but channel 2 from both the ``real data'' and ``simulated data'' were used as independent utterances for pre-training and fine-tuning. Channel 2 was excluded because it is the one corresponding to the ``behind'' microphone which is of low recording quality. Since the CHiME-4 corpus is too small (less than 100 hours) to pre-train from scratch, we continued the pre-training with the CHiME-4 corpus on top of a model already pre-trained with LibriSpeech. The detailed comparisons between with and without continual pre-training are available in Table \ref{tab:chime}.

\vspace{-2mm}
\subsection{Model Pre-training}
Our implementation was made upon the official release of wav2vec 2.0  from \fairseq \cite{ott2019fairseq}, and the network architecture used throughout this paper was identical to the LibriSpeech \textsc{Base}\xspace configuration specified in \cite{baevski2020wav2vec}:  12 transformer blocks with hidden dimension 768 and 8 heads. Most of the training settings and strategies for the \textsc{Base}\xspace model were also carried over into ours, except that the batch sizes were doubled to accommodate the pairs of original-noisy examples. The coefficient $\lambda$ in Eq. (\ref{eq:new_contrastive_loss}) was set to 0.3 empirically for all the wav2vec-Switch experiments. We also found that a smaller learning rate (e.g., 1/5 of the one used in the pre-training with LibriSpeech) for the continual training led to better ASR performance.

All the models were trained with 32 NVIDIA Tesla V100 GPUs, each with 32GB memory (required for the double-sized batches). We picked the best checkpoint (in terms of the validation loss), rather than the last one, for continual pre-training or fine-tuning after that, so that we do not need to worry about the risk of over-training.

\vspace{-2mm}
\subsection{Model Fine-tuning}
We followed the \texttt{base\_100h} setup in the wav2vec 2.0 code except the specific data being used for the CHiME-4 experiments. All the fine-tuning models were trained with 2 GPUs. We chose the best checkpoint according to the validation WER for final evaluations.

\vspace{-2mm}
\subsection{Decoding and Language Models}
As we employed the CTC loss for fine-tuning, incorporating a language model during decoding is crucial for the best performance. For the LibriSpeech experiments, we downloaded an existing Transformer word-level language model\footnote{\url{https://dl.fbaipublicfiles.com/wav2letter/sota/2019/lm/lm_librispeech_word_transformer.pt}} and adopted the same decoding configurations as those corresponding to the \textsc{Base}\xspace 100-hour experiment introduced in the wav2vec 2.0 paper. For the CHiME-4 experiments, an LSTM-based word-level language model with a vocabulary size of 65,000 was trained on the text portion of the WSJ corpus (see \cite{wang2019espresso} for model details), and then the same decoding strategy was performed. The LM weight was tuned separately for each model.

\vspace{-1mm}
\subsection{Results on Synthesized Noisy Data}
\label{sec:synthesized}
We first show the WER (\%) results on both the original and synthesized noisy sets under the matched condition in Table \ref{tab:librispeech}. Besides wav2vec-Switch (the third row), we also include the results from the wav2vec 2.0 model pre-trained on the original 960-hour LibriSpeech corpus (the first row), and the model trained on the synthesized noisy data (the second row, a.k.a. the data augmentation baseline). It is not surprising that without ``seeing'' the noisy data in training, the performance on the noisy test sets is much worse than that on the original ones: WERs increase by 2.5--3.2 times. After adding noise to the training data, the performance on the noisy data gets greatly improved, while not changing significantly on the original sets. When further replacing the wav2vec 2.0 model with wav2vec-Switch, the performance on the noisy sets is improved relatively by 7.1--11.0\% without LM, or 2.9--4.9\% with LM. In addition, the WERs on the original sets are even slightly better than those in the baseline in the ``no LM'' case, and remain almost the same if the LM is used. These results are promising as it is shown clearly that the model trained with wav2vec-Switch achieves noise robustness on the synthesized noisy data without hurting the performance on the original data.

\vspace{-1mm}
\begin{table}[ht]
\caption{Results on the original and synthesized noisy LibriSpeech test sets under the matched condition.}
\vspace{-4mm}
\begin{center}
\begin{adjustbox}{max width=\linewidth}
\begin{tabular}{lccccc}
\toprule
  & \multirow{2}{*}{\textbf{LM}}   & \multicolumn{2}{c}{\textbf{Original}} & \multicolumn{2}{c}{\textbf{Noisy} (5--10 dB)} \\
\cmidrule(lr){3-4} \cmidrule(lr){5-6}
                &                & test-clean  & test-other  & test-clean  & test-other  \\ \midrule
\multirow{2}{*}{wav2vec 2.0}        & N         &   5.9       & \textbf{13.4}        & 15.6        & 33.1        \\
                                              & Y         &   \textbf{2.6}       & \textbf{6.6}         & 8.0        & 21.3        \\ \midrule
\multirow{2}{*}{\makecell{\quad+ MUSAN \emph{music}+\emph{noise} (5--10 dB)\\ (Baseline)}}  & N         &   6.1       & 14.1        & 8.2      & 19.8        \\
                                              & Y         &   \textbf{2.6}       & 6.7        & 3.4         & 10.2        \\ \midrule
\multirow{2}{*}{wav2vec-Switch}              & N         &   \textbf{5.8}       & 13.6        & \textbf{7.3}         & \textbf{18.4}        \\ 
                                              & Y         &   2.7      & 6.7       & \textbf{3.3}      & \textbf{9.7}        \\ \bottomrule
\end{tabular}
\end{adjustbox}
\end{center}
\label{tab:librispeech}
\vspace{-2mm}
\end{table}
\vspace{-3mm}

Next we present the results under mismatched conditions. \emph{Mismatched conditions} refers to the cases where the noisy conditions for testing are different from those for training. Given that the noisy condition for training was \emph{music}+\emph{noise} (5--10 dB), we created 3 versions of noisy test sets with mismatched conditions: 1) \emph{music}+\emph{noise} (0--5 dB) was the version where the SNR range in the test sets has no overlap with that in the training set; 2) \emph{speech} (5--10 dB) was the one where the noise type being added to the test sets is different; and 3) \emph{speech} (0--5 dB) was when both the SNR range and noise type are different. Table \ref{tab:librispeech_mismatch} demonstrates the results without LM along with the gains obtained by using the proposed wav2vec-Switch instead of wav2vec 2.0. We can see that the improvements hold in all the mismatched conditions. However, the relative gain becomes smaller when the degree to which the test noise condition mismatch with the training noise condition gets larger.

\begin{table}[ht]
\caption{Results on the synthesized noisy sets under different mismatched noisy conditions (without LM).}
\vspace{-4mm}
\begin{center}
\begin{adjustbox}{max width=\linewidth}
\begin{tabular}{lcccccc}
\toprule
        & \multicolumn{2}{c}{\emph{music}+\emph{noise} (0--5 dB)} & \multicolumn{2}{c}{\emph{speech} (5--10 dB)} & \multicolumn{2}{c}{\emph{speech} (0--5 dB)}\\
\cmidrule(lr){2-3} \cmidrule(lr){4-5} \cmidrule(lr){6-7}
                                    & test-clean  & test-other  & test-clean  & test-other  & test-clean  & test-other \\ \midrule
\makecell{wav2vec 2.0 + MUSAN \\\emph{music}+\emph{noise} (5--10 dB)}   &   11.0      & 26.1        & 25.7      & 52.9  & 54.7 & 82.4  \\ 
wav2vec-Switch        &   \textbf{9.7}      & \textbf{24.5}   & \textbf{23.8}     & \textbf{50.4}   &  \textbf{52.7}  & \textbf{80.7}  \\ \midrule
Gain (\%)             &     11.8            & 6.1             & 7.4               & 4.7             & 3.7             & 2.1 \\
\bottomrule
\end{tabular}
\end{adjustbox}
\end{center}
\label{tab:librispeech_mismatch}
\end{table}

\vspace{-7mm}
\subsection{Results on Real Noisy Data}
\label{sec:real}
We now evaluate wav2vec-Switch on the CHiME-4 corpus. The 1-channel track real noisy data was used for model validation and evaluation. We also include the best results reported in the challenge \cite{du2016ustc} and other more recent ones \cite{chen2018building,wang2020complex}. All of them adopted a supervised training paradigm and had a dedicated speech enhancement module to preprocess the noisy speech input before an acoustic modeling module, and some of them even leveraged model ensemble. As presented in Table \ref{tab:chime}, the best results are from wav2vec-Switch with the continual pre-training. The relative improvement from the corresponding baseline is 7.8\% without LM (16.5 vs. 17.9), or 5.7\% with LM (6.6 vs. 7.0). It is also worth noting that, without any speech enhancement, our self-supervised approach followed by a simple CTC fine-tuning achieves better results than those using carefully designed enhancement algorithms. The additional data we were using was just the unlabeled 960-hour LibriSpeech audio and the MUSAN corpus. 

\begin{table}[ht]
\caption{Results on the CHiME-4 real 1-channel dev/eval sets.}
\vspace{-5mm}
\begin{center}
\begin{adjustbox}{max width=\linewidth}
\begin{tabular}{lcccc}
\toprule
  & \textbf{continual pre-training} & \textbf{LM} & dev & eval \\
\midrule
Chen et al. (Kaldi Baseline) \cite{chen2018building} (2018)           & & Y         &   5.6       & 11.4 \\
Du et al. \cite{du2016ustc} (2016)                                    & & Y         &   4.6       & 9.2  \\
Wang et al. \cite{wang2020complex} (2020)                             & & Y         &   \textbf{3.5}       & 6.8  \\
\midrule
\multirow{4}{*}{\makecell{wav2vec 2.0 + MUSAN \emph{music}+\emph{noise} (5--10 dB)\\ (Baseline)}} & \multirow{2}{*}{N} & N     &   10.6       & 17.6      \\
                                              & & Y         &   3.7       & 7.2        \\ \cmidrule(lr){2-5}
                                              & \multirow{2}{*}{Y} & N     &   10.7       & 17.9      \\
                                              & & Y         &   4.6       & 7.0        \\\midrule
\multirow{4}{*}{wav2vec-Switch}    & \multirow{2}{*}{N} & N         &   10.2       & 16.8    \\ 
                                             & & Y         &   3.6      & 7.1        \\ \cmidrule(lr){2-5}
                                   & \multirow{2}{*}{Y} & N         &   \textbf{10.0}       & \textbf{16.5}    \\
                                             & & Y         &   \textbf{3.5}      & \textbf{6.6}        \\\bottomrule
\end{tabular}
\end{adjustbox}
\end{center}
\label{tab:chime}
\vspace{-6mm}
\end{table}

\vspace{-3mm}
\subsection{Ablation Study}
\label{sec:ablation}
To investigate the impact of not keeping the dropout masks or the masked positions the same within each original-noisy speech pair, we conducted one experiment on LibriSpeech where the dropout masks within the context network were different between the examples within each pair, and another experiment where the masked positions were different\footnote{We still keep the number of masked positions the same to ensure dimension match after switching the targets.}. We report the results without LM in Table \ref{tab:ablation}. For easy comparisons, we also copy 2 relevant rows from Table \ref{tab:librispeech}. It shows that despite a clear degradation from the one with identical dropout masks, the one with non-identical dropout masks is still better than the baseline on all the test sets except the original \texttt{test-clean}. However, having different masked positions resulted in a significant deterioration in WER (15.8--27.6\% increase), which is expected since we cannot force a representation to still have consistent predictions when they are actually making predictions for different positions before and after the switch, and consequently it will end up with a sub-optimal solution if doing so. These two experiments attest the importance of maintaining identical dropout masks and masked positions for input pairs in wav2vec-Switch. 

\begin{table}[ht]
\caption{Results on the original and synthesized noisy LibriSpeech sets related to whether to keep identical dropout masks or masked positions between speech pairs. The models were evaluated without LM. Row 1 and 2 are from Table \ref{tab:librispeech} for clearer comparisons.}
\vspace{-4mm}
\begin{center}
\begin{adjustbox}{max width=\linewidth}
\begin{tabular}{lccccc}
\toprule
    & \multicolumn{2}{c}{\textbf{Original}} & \multicolumn{2}{c}{\textbf{Noisy} (5--10 dB)} \\
\cmidrule(lr){2-3} \cmidrule(lr){4-5}
                              & test-clean  & test-other  & test-clean  & test-other  \\ \midrule
\makecell{wav2vec 2.0 + MUSAN \emph{music}+\emph{noise} (5--10 dB) \\ (Baseline)} &   6.1       & 14.1        & 8.2         & 19.8    \\ \midrule
wav2vec-Switch w/ identical dropout masks &   \textbf{5.8}       & \textbf{13.6}        & \textbf{7.3}         & \textbf{18.4}    \\\midrule
wav2vec-Switch w/o identical dropout masks &   6.4       & 13.8        & 7.8         & \textbf{18.4} \\ \midrule
wav2vec-Switch w/o identical masked positions &  7.4      & 16.5     &  8.6     & 21.3 \\ \bottomrule
\end{tabular}
\end{adjustbox}
\end{center}
\label{tab:ablation}
\vspace{-2mm}
\end{table}

\vspace{-5mm}
\section{Conclusion}
\vspace{-1mm}
We present wav2vec-Switch, a self-supervised learning model based on wav2vec 2.0 that infuses noise robustness into the contrastive representation learning for ASR without introducing extra components to neural networks. Robustness to noise is achieved by feeding pairs of original and noisy speech into the network, and enforcing the prediction consistency for the original and noisy speech, i.e. treating the corresponding quantized targets of the original and noisy speech as positive in the contrastive learning. Experiments on both synthesized and real-world noisy speech exhibit the power of our proposed method for robust ASR. Future work includes pre-training on even larger unlabeled data, exploring other ways of getting contrastive samples, and extending our method beyond the contrastive learning.

\vfill\pagebreak

\bibliographystyle{IEEEbib}
\fontsize{8.7}{10.1}\selectfont
\bibliography{strings,refs}

\end{document}